\documentclass[sigconf, screen]{acmart}

\usepackage{url}
\usepackage{longtable}
\usepackage{makecell}
\usepackage{fontawesome}
\usepackage{setspace}
\usepackage{graphicx}
\usepackage{booktabs}
\usepackage{amsmath}
\usepackage{multirow}
\usepackage{multicol}
\usepackage{tabularx}
\usepackage{enumitem}
\usepackage{collcell}
\usepackage{color, colortbl}
\usepackage{cleveref}
\usepackage{makecell}
\usepackage{arydshln}
\usepackage{array}
\usepackage{cleveref}
\usepackage{multibib}
\usepackage{color,soul}
\usepackage{geometry}

\newcolumntype{H}{>{\setbox0=\hbox\bgroup}c<{\egroup}@{}}

\graphicspath{{images/}}

\newcommand{\Lagr}{\mathcal{L}}

\newcommand{\ie}{i.e.,~}
\newcommand{\eg}{e.g.,~}

\newcommand{\Ni}{({\em i})~}
\newcommand{\Nii}{({\em ii})~}
\newcommand{\Niii}{({\em iii})~}

\definecolor{cyan}{rgb}{0.88,1,1}

\newcommand{\modf}[1]{\textcolor{black}{#1}}

\newcommand{\mystar}{{\fontfamily{lmr}\selectfont$\star$}}

\copyrightyear{2022}
\acmYear{2022}
\setcopyright{rightsretained}

\acmConference[KDD '22]{Proceedings of the 28th ACM SIGKDD Conference on Knowledge Discovery and Data Mining}{August 14--18, 2022}{Washington, DC, USA}
\acmBooktitle{Proceedings of the 28th ACM SIGKDD Conference on Knowledge Discovery and Data Mining (KDD '22), August 14--18, 2022, Washington, DC, USA}
\acmISBN{978-1-4503-9385-0/22/08}
\acmDOI{10.1145/3534678.3539169}

\usepackage{etoolbox}
\makeatletter
\patchcmd{\maketitle}{\@copyrightpermission}{
   \begin{minipage}{0.3\columnwidth}
     \href{https://creativecommons.org/licenses/by-nc-nd/4.0/}{\includegraphics[width=0.90\textwidth]{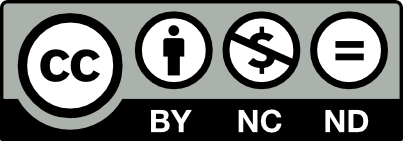}}
   \end{minipage}\hfill
   \begin{minipage}{0.7\columnwidth}
     \href{https://creativecommons.org/licenses/by-nc-nd/4.0/}{This work is licensed under a Creative Commons Attribution-NonCommercial-NoDerivs International 4.0 License.}
   \end{minipage}

   \vspace{5pt}
}{}{}

\makeatother

\begin{document}

\title{Preventing Catastrophic Forgetting in Continual Learning of New Natural Language Tasks}

\author{Sudipta Kar}
\authornote{Work done during internship.}
\affiliation{%
  \institution{Amazon}
  \city{Seattle, WA}
  \country{USA}}
\email{sudipkar@amazon.com}

\author{Giuseppe Castellucci}
\affiliation{%
  \institution{Amazon}
  \city{Seattle, WA}
  \country{USA}}
\email{giusecas@amazon.com}

\author{Simone Filice}
\affiliation{%
 \institution{Amazon}
 \city{Tel Aviv}
 \country{Israel}}
 \email{filicesf@amazon.com}

\author{Shervin Malmasi}
\affiliation{%
  \institution{Amazon}
  \city{Seattle, WA}
  \country{USA}}
  \email{malmasi@amazon.com}

\author{Oleg Rokhlenko}
\affiliation{%
  \institution{Amazon}
  \city{Seattle, WA}
  \country{USA}}
  \email{olegro@amazon.com}

\renewcommand{\shortauthors}{Sudipta Kar et al.}

\begin{abstract}
Multi-Task Learning (MTL) is widely-accepted in Natural Language Processing as a standard technique for learning multiple related tasks in one model. Training an MTL model requires having the training data for all tasks available at the same time. As systems usually evolve over time, (e.g., to support new functionalities), adding a new task to an existing MTL model usually requires retraining the model from scratch on all the tasks and this can be time-consuming and computationally expensive. Moreover, in some scenarios, the data used to train the original training may be no longer available, for example, due to storage or privacy concerns.

In this paper, we approach the problem of incrementally expanding MTL models' capability to solve new tasks over time by \emph{distilling} the knowledge of an already trained model on $n$ tasks into a new one for solving $n+1$ tasks. 
To avoid catastrophic forgetting, we propose to exploit unlabeled data from the same distributions of the old tasks. Our experiments on publicly available benchmarks show that such a technique dramatically benefits the distillation by preserving the already acquired knowledge (\ie preventing up to 20\% performance drops on old tasks) while obtaining good performance on the incrementally added tasks. Further, we also show that our approach is beneficial in practical settings by using data from a leading voice assistant.

\end{abstract}



\maketitle

\section{Introduction}\label{sec:intro}
In recent years, voice assistants, like Alexa or Siri, have become very popular. They exploit Natural Language Processing (NLP) capabilities (\eg intent classification, slot filling, etc.) to provide useful functionalities to the users, \modf{\eg asking questions, getting the latest news, getting weather updates, etc.}
Multi-Task Learning (MTL) models are often preferred in such systems as they have multiple advantages: \Ni MTL normally leads to a better generalization of the model by exploiting the domain-specific information in the training signals of the related tasks \cite{ruder2017}; \Nii MTL models are easier to deploy and maintain when compared to multiple single task models; \Niii MTL models can reduce overall inference latency as they solve multiple tasks in a single inference step.

Modern systems are continually updated over time to support new functionalities,  \eg new intents and slots, or even new tasks.
In MTL, accommodating new tasks usually means training a new model from scratch by using all the past and new training data for all the tasks to be supported. This process can be both time-consuming and computationally expensive. Moreover, this may not always be possible in practical settings as the past data might no longer be available. There may be storage, regulatory, customer, business, or privacy-related constraints, as well as issues of missing and corrupted data. In such cases, the original training data cannot be used, making a full re-training of the MTL model infeasible.

Continual Learning (CL) proposes a viable solution to enable models to keep on learning over time \cite{lifelongLiu}. In CL, a model is updated by only considering the newest task's training data. Thus, a major concern in CL is that it can cause the so-called Catastrophic Forgetting (CF) \cite{mccloskey1989,goodfellow2013empirical}, where performance drops in the previously learned tasks due to the different data distribution of the new training set \cite{Lesort2021UnderstandingCL}.
A solution to this problem is to use Knowledge Distillation (KD) \cite{kd_hinton}, to ``distill'' (i.e., transfer\footnote{Transfer learning usually refers to using a model trained on a source domain to help target domains learning, but it is not continual and it has no knowledge retention mechanism.}) the knowledge from a teacher model to a student model. 
The student learns a task from the soft targets (i.e., output scores) produced by the teacher. In this way, the new model can exploit the teacher's uncertainty contained in its scores as well. KD is often applied for model compression to shrink a large model into a smaller but similarly performing one.
In Computer Vision, KD has been successfully adopted in a CL setting (\eg Li and Hoiem \cite{lihoiem2017}) where a model acts as the teacher providing soft targets for $n$ tasks to a student model that is trained to solve the same set of tasks plus a new one. Applying such a technique in NLP for continual learning is not straightforward, as the data distribution for the different tasks can vary greatly. For example, let us assume we have an intent classification model trained to support an initial set of intents. If we need to extend the model to support a very different set of intents, we might expect a drastic change in the incoming utterance distribution. This can negatively impact the application of the standard KD technique.

We propose a semi-supervised continual learning solution for extending a model already trained on $n$ tasks to solve also a new one, by using only the new tasks' training data as annotated material.
In order to prevent the catastrophic forgetting phenomenon, we exploit a set of \textit{unlabeled} material from the data distributions of the existing tasks. In particular, we aim to distill the old model's (the teacher) knowledge to a new one (the student) by using such unlabeled data, which is supposed to better resemble the old training material data distribution.

We experiment with our approach on a set of text classification tasks from the GLUE benchmark \cite{glue} using BERT-based models. We compare different continual learning strategies, ranging from simple re-training to traditional KD under the assumption that the training data of the previous tasks is no longer available. Experimental results show that our proposed approach can effectively learn sequences of tasks, where at each stage there is no, or very little, forgetting of the previous tasks (\ie preventing up to 20\% performance drops on old tasks) while still being capable of learning to solve a new task.
Moreover, we perform similar experiments in a real setting by exploring a dataset collected from a real voice assistant. We show that our proposed approach allows updating an existing MTL model for the Intent Classification task on different domains while preventing catastrophic forgetting.
Our contributions can be summarized as follows: 
\begin{itemize}
    \item We show that we can  minimize the catastrophic forgetting phenomenon during incremental learning of NLP tasks.
    \item With our CL approach, we achieve comparable results with fully retrained MTL models.
    \item We show that the general knowledge of pre-trained language models is preserved even after several fine-tuning phases.
    \item Our approach works both on publicly available benchmarks (\ie GLUE) and real scenarios over real voice assistant data.
\end{itemize}

The rest of the paper is organized as follows: Section \ref{sec:related} discusses the related works; Sections \ref{sec:methods} and \ref{sec:experiments} present the continual learning approach and the experiments, respectively. Finally, Section \ref{sec:conclusions} discusses our conclusions and future work.

\section{Related Work}\label{sec:related}
Multi-Task Learning (MTL) \cite{Caruana1993MultitaskLA} models achieve impressive results in several domains \cite{ruder2017}, including Natural Language Understanding \cite{clark-etal-2019-bam,liu-etal-2019-multi-task,pentyala-etal-2019-multi}. 
However, MTL models require to be trained from scratch every time there is the need to adapt them to new tasks, which can be infeasible in many scenarios. In contrast, Continual Learning (CL) (also referred to as lifelong learning \cite{lifelongLiu}) studies the problem of learning from a stream of data and extending the acquired knowledge \cite{lange2019continual} over time. The stream can change during time by an evolution in the input distribution or by incorporating new tasks. A key challenge in adding new tasks with CL strategies is to avoid the Catastrophic Forgetting (CF) \cite{goodfellow2013empirical} phenomenon.

CL has been explored mostly in Computer Vision \cite{8237630,lihoiem2017,Rannen_2017_ICCV,Oswald2020Continual,NEURIPS2020_d7488039,9577808,He2021UnsupervisedCL,rebuffi-cvpr2017}.  \modf{Few works involve continual learning in the NLP domain. In \cite{pgn_rusu2016progressive} the authors introduce the PGN architecture where new copies of a model are issued for each new task. Elastic Weight Consolidation \cite{ewc_Kirkpatrick3521} is another CL technique, where task-specific constraints on the model weights are added to prevent the catastrophic forgetting phenomenon. In the slot filling task, the work of \citet{shen-etal-2019-progressive} proposes a strategy to progressively increase the slot types the model can recognize. In \cite{liu-etal-2019-continual}, the authors train sentence encoders with unsupervised methods to continually learn features from new corpora. Similarly, \citet{Jin2021LifelongPC} discuss the problem of continually adapting large pre-trained language models to new settings. In \cite{Li2020Compositional}, the authors address the open and growing vocabulary problem in a sequence-to-sequence framework. \citet{huang-etal-2021-continual} propose a regularization approach based on information disentanglement by separating the hidden space in task general and task-specific sub-spaces by utilizing auxiliary tasks.}

Most of these works are either not directly applicable to the setting proposed here or are unfeasible with large pre-trained models. In \cite{Monaikul2021ContinualLF} a CL approach using a KD mechanism is proposed for Named Entity Recognition problems, where the authors incrementally train a sequence tagging model to support new entities over time. \citet{castellucci-etal-2021-learning} used distillation to continually adapt to new languages. In this work, instead, we aim to adopt an already trained model to new NLP tasks with large pre-trained models like BERT by exploiting KD in a teacher-student setting \cite{8237630}, where we make use of unlabeled text data to prevent forgetting of the previously learned tasks.

\section{Continual Learning for Natural Language  Tasks}\label{sec:methods}

\begin{figure*}[!ht]
    \centering
    \includegraphics[width=0.90\textwidth]{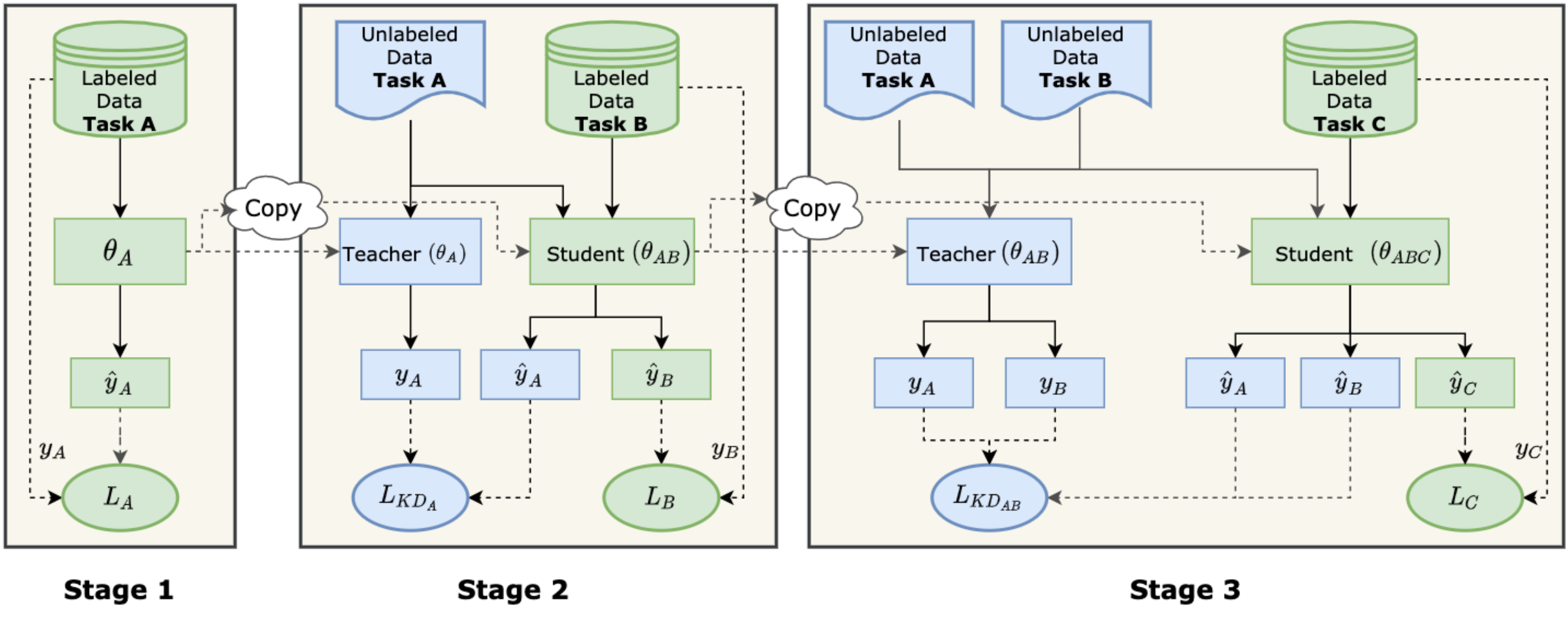}
    \caption{An illustration of incrementally learning three tasks through our proposed CL approach. \textbf{Stage 1:} A model $\theta_A$ is trained on \texttt{Task A} using a set of labeled data and by minimizing the Cross Entropy (CE) loss $\Lagr_{A}$. Here $y_A$ indicates the ground truth labels for \texttt{Task A} from the labeled dataset, and $\hat{y}_A$ indicates the predictions for \texttt{Task A} by $\theta_A$. \textbf{Stage 2:} The trained $\theta_A$ model from Stage 1 is treated as a frozen \texttt{Teacher} model. A trainable \texttt{Student} copy $\theta_{AB}$ is created by adding a new classification head for \texttt{Task B}. The \texttt{Student} model is trained on the CE based task loss $\Lagr_{B}$ for \texttt{Task B} and distillation loss $\Lagr_{KD_{A}}$ for \texttt{Task A}. To compute $\Lagr_{KD_{A}}$, a set of unlabeled data is used: \texttt{Teacher} model's predictions on such dataset for \texttt{Task A} are treated as soft labels and are used against the \texttt{Student} model's predictions. In this way, the \texttt{Student} learns to perform \texttt{Task B} and at the same time tries to keep \texttt{Task A}'s knowledge by distilling it from the \texttt{Teacher} model. \textbf{Stage 3:} The \texttt{Student} from stage 2 $\theta_{AB}$ acts as the frozen \texttt{Teacher}, and a \texttt{Student} copy $\theta_{ABC}$ is created to add \texttt{Task C}. The rest of the training process is similar to stage 2.}
    \label{fig:cl_nlu_schema}
\end{figure*}

To continually learn new tasks for NLP, we exploit the KD \cite{kd_hinton} framework. Without loss of generality, let us assume that we have already trained a model $\theta_A$ to solve task $A$ and we want to update it to learn how to also solve a new task B. As illustrated in Figure \ref{fig:cl_nlu_schema}, we start by creating a copy of $\theta_A$, by adding a new output layer for task B, i.e., $\theta_{AB}$. The original $\theta_A$ and $\theta_{AB}$ models act as the teacher and the student in the KD framework, respectively.
During training, we keep we $\theta_A$ frozen and we only update $\theta_{AB}$ with the objective of \Ni learning the new task $B$ from the training data and \Nii preserving the older task's knowledge by minimizing the loss function $$\Lagr_{AB} = \Lagr_{KD_A} + \Lagr_{B}.$$
Let us consider a set of training examples $(\boldsymbol{x},\boldsymbol{y})$, where $\boldsymbol{x}$ is an input representation and $\boldsymbol{y}$ is a target category. The distillation loss is defined as $\Lagr_{KD_A}= CE(\theta_{A}(\boldsymbol{x}), \theta_{AB}(\boldsymbol{x}))$, i.e., the cross-entropy (CE) between the class probability distribution of the student on task $A$ and the soft targets derived from the teacher $\theta_A$ by applying a temperature-controlled softmax. The temperature-controlled softmax converts the logit $z_i$ of each class into a probability $$q_i = \frac{exp(z_i/T)}{\sum_j exp(z_j/T)}$$ where $T$ controls the smoothness of the output distribution \cite{kd_hinton}. Using higher values of $T$ will make the output distribution softer, and can be used to control the influence of the soft targets on the loss. Simultaneously, the model is minimizing also the new task loss, which is defined as $\Lagr_{B} = CE(\boldsymbol{y}, \theta_{AB}(\boldsymbol{x}))$, i.e., the usual CE with respect to the annotated targets for task B.
While $\Lagr_{B}$ serves to let the student learn how to solve a new task, $\Lagr_{KD_A}$ helps it in preventing catastrophic forgetting of the old one. In the standard application of KD to CL, $\Lagr_{KD_A}$ is computed on the new task data: this assumes that the old and new tasks have the same data distribution. For instance, in Computer Vision, when models are trained to recognize an increasing number of object classes, the input images come from the same underlying distribution \cite{8237630}.

However, in NLP, tasks are typically defined on very different data distributions, and preventing the catastrophic forgetting when using only the new task data can be challenging. As an example, let us assume we want to add the answer selection task to a model trained for paraphrase identification. In this case, examples for the new task are $($\textit{question}, \textit{candidate\_answer}$)$  pairs, which, intuitively, are not paraphrases. If we use this data to compute the distillation loss we never consolidate the model capability to recognize paraphrases but we bias it toward the non-paraphrase class.

\subsection{Dealing with Different Data Distributions}\label{sec:ukd}

We propose augmenting the KD learning process with a data distribution resembling the one used to train the teacher model to solve task A. Our assumption is that while the original training material for task $A$ may no longer be available, we can still observe a stream of unlabeled data ($U_A$) from the same distribution. For example, the raw data on which the teacher model is applied for inference. Note that in case of storage constraints we can still process the data on the fly to accumulate gradients and run mini-batch training.

By doing so, the loss function $\Lagr_{KD_A}$ represents the discrepancy between the teacher and student predictions for the old tasks on a set of unlabeled data. In practice, the unlabeled data $U_A$ are automatically labeled by the teacher model $\theta_{A}$ to produce the soft targets dataset of  task A. This dataset will be used to compute the loss $\Lagr_{KD_A}$. Instead, a new labeled  dataset for task $B$ is used to compute $\Lagr_{B}$. In this way, the student model should be able to minimize the discrepancy with the teacher on the old task (\ie minimizing the catastrophic forgetting) while learning the new task.

This methodology can be easily extended to the general case where the teacher is already trained on $n$ tasks and the student needs to solve a new task. In this setting, we need to prevent the catastrophic forgetting of $n$ different tasks. We assume the availability of an unlabeled stream of data for each of the old tasks in order to compute the individual distillation losses. In this way, the student model will maintain the relevant knowledge to solve the $n$ tasks by distilling it from the teacher on the unlabeled data stream, while also learning how to solve the new task on the labeled data. We will refer to this proposed approach of using unlabeled data for knowledge distillation as UKD throughout the rest of the paper. 

\section{Experiments and Results}\label{sec:experiments}
\subsection{Dataset}\label{sec:dataset}
\begin{table}[t]
    \centering
    \resizebox{\columnwidth}{!}{
    \begin{tabular}{ p{\columnwidth}}
    \toprule
        \textbf{Dataset:} Multi-genre NLI Corpus (MNLI)  \\
         \textbf{Train-Dev-Test Size:} 314k-79k-20k\\
         \textbf{Evaluation Metric:} Accuracy for \textit{matched} and \textit{mismatched} \\
         \textbf{Example:}\\
            \textit{\textbf{S1}: i don't know um do you do a lot of camping}\\
            \textit{\textbf{S2}: I know exactly.}\\
            \textit{\textbf{Label:} Contradiction}\\ \midrule
    
         \textbf{Dataset:} Quora Question Pairs (QQP)  \\
         \textbf{Train-Dev-Test Size:} 291k-73k-391k\\
         \textbf{Evaluation Metric:} F1/ Accuracy \\
         \textbf{Example:}\\
            \textit{\textbf{Q1:} What are the best things to do in Hong Kong?}\\
            \textit{\textbf{Q2:} What is the best thing in Hong Kong?}\\
            \textit{\textbf{Label:} Duplicate}\\ \midrule
        
        \textbf{Dataset:} Question NLI (QNLI)  \\
         \textbf{Train-Dev-Test Size:} 84k-21k-5.4k\\
         \textbf{Evaluation Metric:} Accuracy \\
         \textbf{Example:}\\
            \textit{\textbf{Q:}What language did Tesla study while in school?}\\
            \textit{\textbf{A:} Tesla was the fourth of five children.}\\
            \textit{\textbf{Label:} Not entailment}\\ \midrule
    
        \textbf{Dataset:} Mircrosoft Research Paraphrase Corpus (MRPC)  \\
         \textbf{Train-Dev-Test Size:} 2.9k-740-1.7k\\
         \textbf{Evaluation Metric:} F1/ Accuracy \\
         \textbf{Example:}\\
            \textit{\textbf{S1:} Amrozi accused his brother, whom he called "the witness" , of deliberately distorting his evidence.}\\
            \textit{\textbf{S2:} Referring to him as only "the witness", Amrozi accused his brother of deliberately distorting his evidence.}\\
            \textit{\textbf{Label:} Paraphrase}\\ \midrule
            
        \textbf{Dataset:} Recognizing Textual Entailment (RTE)  \\
         \textbf{Train-Dev-Test Size:} 2k-500-3k\\
         \textbf{Evaluation Metric:} Accuracy \\
         \textbf{Example:}\\
            \textit{\textbf{S1:} No Weapons of Mass Destruction Found in Iraq Yet.}\\
            \textit{\textbf{S2:} Weapons of Mass Destruction Found in Iraq.}\\
            \textit{\textbf{Label:} Not Entailment}\\ \midrule
        
        \textbf{Dataset:} Stanford Sentiment Treebank (SST-2)  \\
         \textbf{Train-Dev-Test Size:} 53.6k-13.4k-1.8k\\
         \textbf{Evaluation Metric:} Accuracy \\
         \textbf{Example:}\\
            \textit{gorgeous and deceptively minimalist }\\
            \textit{\textbf{Label:} Positive}\\ 
    \bottomrule
    \end{tabular}}
    \caption{Details about the tasks and their dataset. We treat the official development set as our test set and use 20\% of the training set as our development set.}
    \label{tab:dataset}
\end{table}

We conduct our continual learning experiments on the following datasets, which are part of the GLUE benchmark \cite{glue}: Multi-genre Natural Language Inference (MNLI; \citeauthor{MNLI}, \citeyear{MNLI}), Quora Question Pairs (QQP; \citeauthor{QQP}, \citeyear{QQP}), Microsoft Research Paraphrase Corpus (MRPC; \citeauthor{MRPC}, \citeyear{MRPC}),  Question Natural Language Inference (QNLI), Recognizing Textual Entailment (RTE), and Stanford Sentiment Treebank (SST-2; \citeauthor{sst2}, \citeyear{sst2}). These represent a diverse set of NLP text classification tasks with different data distributions, \ie we can expect to observe the catastrophic forgetting phenomenon when updating an already trained model. In Table \ref{tab:dataset} we provide an overview of the tasks and their dataset. Notice that these datasets are also characterized by different sizes for the training material. We expect that catastrophic forgetting could be worse for those tasks where the training set is smaller.

We use Accuracy and F1 score to measure performance on different datasets, following the literature.
For MNLI, we report accuracy for both \textit{matched} (genres are the same as the training set) and \textit{mismatched} (genres different than the training data) sets of the data.

\subsection{Experimental Setup}\label{sec:training}
In our experiments, we use the pre-trained BERT base uncased model \cite{devlin-etal-2019-bert} of the Transformers \cite{Wolf2019HuggingFacesTS} library. To fine-tune BERT, we add an output layer (a dense layer followed by Softmax) for each task on top of the \textit{CLS} token. 
We set the batch size to 64, the learning rate to 5e-6, the dropout to 0.1, and the temperature T to 2. To choose these parameters we ran a set of preliminary experiments with a grid search strategy. We finally trained each model for 20 epochs with early stopping with patience=3.
The results reported in the following sections are based on the official validation set of each task. We used 20\% of the training data for validation.

In our experiments we consider the following CL settings as baselines to compare with our proposed approach using unlabeled data for knowledge distillation (UKD), as described in Section \ref{sec:ukd}.

\begin{itemize}[leftmargin=1.3em,topsep=1em]
    \item \textbf{Single Task (ST):} BERT is fine-tuned for only one task.
    
    \item \textbf{Multi Task Learning (MTL):} We fine-tune BERT through multi-task learning for all the tasks' training material. Notice that this setup assumes the availability of all the training material at the same time. Thus, we can consider MTL as an upper bound to the performance.
    
    \item \textbf{Output Layer (OL):} We take the model $\theta_{A}$ trained on task $A$ and add a new classification head for task $B$. Then, we fine-tune only the new task output layer and freeze the rest of the network of the student model $\theta_{AB}$.

    \item \textbf{Entire Model (EM):} We fine-tune the entire model by considering only the new task's loss. That means all of the parameters of BERT and the old tasks' output layer get updated to learn the new task.
    
    \item \textbf{Elastic Weight Consolidation (EWC):} We impose weight constraints on the model in line with the elastic weight consolidation (EWC) framework,\footnote{We adopted the implementation available in  \url{https://github.com/GT-RIPL/Continual-Learning-Benchmark}} as suggested in \citet{ewc_Kirkpatrick3521} while adapting the model to a new task. 
    
    \item \textbf{Traditional Knowledge Distillation (TKD):} We fine-tune the entire model using both the new task and the traditional knowledge distillation \cite{kd_hinton} losses. The distillation loss for the previous tasks is computed on the new task's training data.
\end{itemize}

In the experiments reported in \cref{sec:twotasks} and \cref{sec:threetasks}, at each training step we use the same unlabeled data (our validation set) for distillation. While this violates our assumption of not being able to store data, it allows us to compare with published results using the development set of the GLUE benchmark. We randomly selected 20\% of the training data to use as our validation set.

In \cref{sec:fivetasks} we show a more realistic case where we use a different unlabeled set at each training stage, i.e., simulating the scenario where we are not allowed to store any data. In this case, results are not comparable with literature since we consider as labeled material only a subset of the training material at each training stage.
Finally, in \cref{sec:realw_exp} we report a set of experimental results obtained by applying our methodology on a dataset extracted from a real voice assistant's utterances.

\subsection{Addition of a Second Task}
\label{sec:twotasks}

\begin{table}[t]
\centering

\resizebox{0.95\columnwidth}{!}{
\begin{tabular}{>{\color{black}} l >{\color{black}} c >{\color{black}} c >{\color{black}}c >{\color{black}}c}

\toprule
\textbf{Setting} & \textbf{Avg.} & \textbf{MNLI} & \textbf{QNLI} & \textbf{QQP}\\
& & Acc. & Acc. & F1 / Acc.\\
\midrule
Single Task  & 87.5 & 83.9, 84.2 & 90.9 & 87.7 / 90.9\\ \midrule
MTL \textsubscript{[MNLI, QNLI]}  & 86.0 & 83.7, 83.8 & 90.4 & - \\ \hdashline
OL \textsubscript{[MNLI, QNLI]}  & 80.0 & \textbf{83.9, 84.2} & 71.8 & - \\
EM \textsubscript{[MNLI, QNLI]}  & 79.2 & 73.2, 73.7 & 90.6 & - \\
EWC \textsubscript{[MNLI, QNLI]}  & 73.8 & 65.6, 65.4 & 90.5 & - \\
TKD \textsubscript{[MNLI, QNLI]}  & 84.0 & 80.3, 80.8 & 	\textbf{90.8} & - \\
UKD \textsubscript{[MNLI, QNLI]}   & \textbf{86.1} & 83.5, 84.0 & 	\textbf{90.8} & -  \\ \hdashline
OL \textsubscript{[QNLI, MNLI]}  & 68.1 & 55.5, 58.0 & 90.9 & - \\
EM \textsubscript{[QNLI, MNLI]}  & 73.2 & \textbf{83.4, 84.1} & 52.1 & - \\
EWC \textsubscript{[QNLI, MNLI]}  & 73.5 & 83.7, 83.7 & 53.1 & - \\
TKD \textsubscript{[QNLI, MNLI]}  & 85.5 & 83.1, 83.6 & 89.9 & - \\
UKD \textsubscript{[QNLI, MNLI]}  & \textbf{86.1} & 83.5, 83.9 & 	\textbf{91.0} & - \\ 
\midrule

MTL \textsubscript{[MNLI, QQP]}  & 86.1 & 83.3, 83.2 & - & 87.3 / 90.4\\ \hdashline
OL \textsubscript{[MNLI, QQP]}  & 78.1 & \textbf{83.9, 84.2} & - & 67.3 / 76.8\\
EM \textsubscript{[MNLI, QQP]}  & 79.8 & 70.9, 71.9 & - & 86.6 / 90.0\\
EWC \textsubscript{[MNLI, QQP]}  & 80.0 & 70.8, 71.5 & - & 87.2 / 90.5\\
TKD \textsubscript{[MNLI, QQP]}  & 85.6 & 81.5, 82.0 & - & 	\textbf{87.8 / 90.9}\\
UKD \textsubscript{[MNLI, QQP]}  & \textbf{86.5} & 	83.5, 84.0 & - & \textbf{87.7 / 90.9} \\ \hdashline
OL \textsubscript{[QQP, MNLI]}  & 70.6 & 51.4, 52.3 & - & 87.7 / 90.9 \\
EM \textsubscript{[QQP, MNLI]}  & 81.1 & 83.0, 83.9 & - & 75.4 / 81.9 \\
EWC \textsubscript{[QQP, MNLI]}  & 82.0 & 83.4, 83.8 & - & 77.3 / 83.3 \\
TKD \textsubscript{[QQP, MNLI]}  & 85.1 & 82.5, 82.6 & - & 85.7 / 89.6 \\
UKD \textsubscript{[QQP, MNLI]}  & \textbf{86.0} & 	\textbf{83.1, 83.0} & - & 	87.2 / 90.6  \\ 
\midrule

MTL \textsubscript{[QNLI, QQP]}  & 89.1 & - & 90.6 & 86.7 / 90.0 \\ \hdashline
OL \textsubscript{[QNLI, QQP]}  & 73.7 & - & \textbf{90.9} & 58.6 / 71.7  \\
EM \textsubscript{[QNLI, QQP]}  & 78.7 & - & 59.3 & 86.8 / 90.1  \\
EWC \textsubscript{[QNLI, QQP]}  & 79.5 & - & 61.0 & 87.1 / 90.4  \\
TKD \textsubscript{[QNLI, QQP]}  & 89.3 & - & 89.7 & 	\textbf{87.5 / 90.6}  \\
UKD \textsubscript{[QNLI, QQP]}  & \textbf{89.2} & - & 	90.1 & 87.0 / 90.4 \\ \hdashline
OL \textsubscript{[QQP, QNLI]}  & 83.8 & - & 72.9 & \textbf{87.7 / 90.9} \\ 
EM \textsubscript{[QQP, QNLI]}  & 84.4 & - & 89.5 & 79.0 / 84.8 \\
EWC \textsubscript{[QQP, QNLI]}  & 83.2 & - & 89.8 & 75.8 / 84.0 \\
TKD \textsubscript{[QQP, QNLI]}  & 88.2 & - & 	\textbf{90.0} & 85.3 / 89.4 \\
UKD \textsubscript{[QQP, QNLI]}  & \textbf{89.2} & - & 89.9 & 87.2 / 90.5 \\ 
\midrule

\end{tabular}}
\caption{Comparison of different CL strategies to learn two tasks in different orders. Comparison of single task and multi-task (MTL) performance with different CL strategies. MT: Multi-task, OL: Fine-tune only the output layer, EM: fine-tune entire model, EWC: elastic weight consolidation \cite{ewc_Kirkpatrick3521}, TKD: traditional knowledge distillation, UKD: knowledge distillation with unlabaled data. [MNLI, QNLI] means QNLI is added to an MNLI model. Best results are in bold. Two different accuracy scores are reported for the \textit{matched} and \textit{mismatched} sets of MNLI.}
\label{app:tab:results_2_tasks}
\end{table}

We investigate different techniques for adding a new task to a model already trained on one task. We experimented with the three largest datasets in the GLUE benchmark: MNLI, QNLI, and QQP. We first fine-tune a pre-trained language model on one task. Then, we add another task to the model through the proposed technique without using any training labels for the first task. We present the results in Table \ref{app:tab:results_2_tasks}, where we report also the performance of the baselines.

As hypothesized, when adding QNLI to a model fine-tuned on MNLI, in the OL setting the model is not able to learn the QNLI task by only adjusting the output layer weights (the new task accuracy is about 20 points below the single task setting (ST)). Alternatively, tuning the entire model (EM) causes catastrophic forgetting of the previous tasks, as demonstrated by the drop of about 10 points in accuracy in both the matched and mismatched settings. We note that the same pattern can also be observed for the other task pairs. Traditional Knowledge Distillation (TKD) brings a balance between these two methods as we observe good performances on both the first and second task. However, when the distillation loss is computed on the second task data (TKD row), the performance gap on the old tasks (w.r.t. ST or MTL) persists. For example, when adding QNLI to a model trained on MNLI data, the TKD approach is about 4 points less than the ST and the MTL systems. This is happening to various degrees to all the old tasks in all the pairs.
Imposing constraints on the model weights, like in the EWC approach, is not effective. We argue that this can be caused by the usage of large and complex pre-trained language models. Given that the drop we observe for EWC is generally higher than other baselines, we will not report the EWC results in the following sections.

In sum, computing the distillation loss with our proposed UKD method largely mitigates the catastrophic forgetting issue and the capability of the model to learn the new task.
When adding QNLI to an MNLI-trained model, the drop of the first task after at the second step is only about 0.2\% when we use the MNLI unlabeled development set for distillation (TKD drop is about 3.5\%).  Additionally, QNLI accuracy when added as a new task is comparable with ST. This means that the model is retaining the general linguistic knowledge required to learn new tasks, while also preserving its knowledge on the old task. Moreover, it is worth noticing that UKD performances are comparable with MTL.
We observe a similar trend in the reverse setting, where we add MNLI to a model fine-tuned on QNLI. Finally, this pattern is consistent in other task pairs as well (e.g., adding QQP to MNLI or QNLI).

\begin{table}[t]
\centering

\resizebox{\columnwidth}{!}{
\begin{tabular}{>{\color{black}} l >{\color{black}}c >{\color{black}}c >{\color{black}}c >{\color{black}}c >{\color{black}}c}

\toprule
\textbf{Setting} & \textbf{Avg.} & \textbf{MNLI} & \textbf{QNLI} & \textbf{QQP} & \textbf{SST-2}\\
& & Acc. & Acc. & F1 / Acc. & Acc.\\
\midrule
Single Task  & 88.5 & 83.9, 84.2 & 90.9 & 87.7 / 90.9 & 93.2 \\ \hline 
Multi-Task  & 86.8 & 83.7, 83.7 & 90.5 & 86.3 / 89.6 &  \\ 
\, \, \, + SST-2  & 87.7 & 83.2, 83.3 & 90.8 & 86.8 / 90.2 & 92.0 \\ \midrule

TKD \textsubscript{MNLI + QNLI + QQP}  & 85.1 & 79.7, 80.1 & 87.9  & 87.1 / 90.5 & \\
\, \, \, + SST-2 & 85.0 & 79.3, 79.1 & 86.6 & 85.3 / 88.3  & 91.5 \\
UKD \textsubscript{MNLI + QNLI + QQP}  & \textbf{86.7} & 83.1, 83.3 & 90.2 & 86.7 / 90.0 & \\
\, \, \, + SST-2 & \textbf{87.0} & 82.6, 82.6 & 90.2 & 83.4 / 90.8 & 92.5 \\ \hline

TKD \textsubscript{MNLI+ QQP + QNLI}  & 84.7 & 80.8, 80.7 & 90.2 & 83.1 / 88.5\\
\, \, \, + SST-2  & 80.7 & 76.8, 76.4 & 89.1 & 69.0 / 82.0 & 90.8\\
UKD \textsubscript{MNLI + QQP + QNLI}  & \textbf{86.8} & {83.2, 83.3} & 89.8 & {87.1 / 90.5}\\
\, \, \, + SST-2  & \textbf{87.3} & 83.1, 83.1 & 89.4 & 86.0 / 90.0 & 92.0 \\ \hline

TKD \textsubscript{QQP + MNLI + QNLI}  & 83.8 & 81.0, 81.9 & {90.1} & 79.5 / 86.5\\
\, \, \, + SST-2   & 77.3 & 67.7, 69.1 & 88.2 & 66.6 / 80.8  & 91.5 \\
UKD \textsubscript{QQP + MNLI + QNLI}  & \textbf{86.5} & {82.6, 83.0} & 90.0 & {86.7 / 90.3}   \\
\, \, \, + SST-2 & \textbf{86.6} & 82.3, 81.7 & 89.3 & 85.1 / 89.5 & 92.0 \\ \hline

TKD \textsubscript{QQP + QNLI + MNLI}  & 85.4 & {83.3, 83.2} & 88.1 & 83.9 / 88.4 \\
\, \, \, + SST-2  & 84.2 & 82.1, 82.2 & 87.5 & 76.6 / 84.8 & 92.1 \\
UKD \textsubscript{QQP + QNLI + MNLI}  & \textbf{86.3} & 82.8, 83.3 & {89.1} & {86.5 / 90.0} \\
\, \, \, + SST-2  &\textbf{ 86.5} & 82.2, 82.4 & 88.2 & 84.6 / 89.2 & 92.2 \\\hline

TKD \textsubscript{QNLI + QQP + MNLI}  & 86.0 & 82.4, 83.1 & 88.4 & 86.2 / 89.8 \\
\, \, \, + SST-2 & 85.5 & 74.6, 76.4 & 97.5 & 84.6 / 87.9 & 92.0\\
UKD \textsubscript{QNLI + QQP + MNLI}  & \textbf{86.6} & {83.2, 82.8} & {90.1} & {86.6 / 90.1} \\ 
\, \, \, + SST-2 & \textbf{86.9} & 83.1, 82.5 & 89.9 & 84.6 / 89.3 & 91.7 \\ \hline

TKD \textsubscript{QNLI + MNLI + QQP}  & 85.9 & 81.2, 81.5 & 89.5 & 86.9 / 90.3  \\
\, \, \, + SST-2  & 83.4 & 77.8, 77.6 & 89.1 & 77.7 / 85.7 & 92.2  \\
UKD \textsubscript{QNLI + MNLI + QQP}  & \textbf{87.0} & 83.3, 83.4 & 90.4 & 87.3 / 90.5  \\ 
\, \, \, + SST-2  & \textbf{87.2} & 83.0, 82.6 & 90.2 & 85.9 / 89.9 & 91.7 \\
\bottomrule

\end{tabular}}
\caption{TKD and UKD performances when incrementally learning three and four tasks. T$_1$ + T$_2$ + T$_3$ means pre-trained BERT was fine-tuned for T$_1$ first. Then T$_2$ and T$_3$ were added incrementally. +SST-2 rows show the result after further adding SST-2.}
\label{app:tab:results_3_tasks}
\end{table}

\subsection{Adding Third and Fourth Tasks}\label{sec:threetasks}

We further explore the effectiveness of UKD by incrementally learning three and four tasks, and we report the results with different task orders in Table \ref{app:tab:results_3_tasks}. 
Observing the weak performance of the OL, EM, and EWC systems for learning the second task, we exclude these baselines for the further experiments as adding more tasks can only degrade the performance more.
Results show that UKD is able to provide useful information to retain the knowledge in the model. For instance, when adding MNLI and QNLI to QQP, the F1 score of QQP drops about 8\% with TKD, while using UKD the drop is only about 1\% compared to the single task model. Notice that this pattern is consistent in almost every task combination we experimented with.

To further confirm the capability of the UKD approach to deal with this setting, we add a very different fourth task, i.e., SST-2. The main difference of this task with MNLI, QNLI and QQP stems from the fact that SST-2 is a single sentence classification task while the others are sentence pair tasks.
Again, using UKD minimizes the forgetting (e.g., 2.6\% drop for QQP vs. 21.1\% drop with TKD). Such improvement is due to the usage of the unlabeled data for the old tasks; without such data the model observes a very different data distribution as the SST-2 task is very different from the other three.

\subsection{Incremental Addition of Five Tasks}
\label{sec:fivetasks}

So far, we used identical unlabeled data for UKD at each step. We now simulate the scenario in which we cannot store any data. We report a continual learning experiment on five different tasks with the order\footnote{Tasks order reflects the inverse order of the dataset sizes to have a reasonable amount of training/distillation data at each step.} MNLI $\rightarrow$ QQP $\rightarrow$ QNLI $\rightarrow$ MRPC $\rightarrow$ RTE, where the unlabeled data is different at every step of UKD.

We divide each training set in equal slices and we use one as labeled data and the others as new unlabeled data in future training steps. For example, the first task (MNLI) is divided in five slices, the second task (QQP) in four slices, and so on. At each step, we train the model on one of the slices of the task we want to add, and we employ an unused slice of each previous task as unlabeled data. For instance, in the first step we simply train a model on one slice of MNLI; in the second stage, we use a different slice of MNLI as unlabeled data for the UKD approach and the first slice of QQP as annotated data to learn the new task, and so on. At the end, we'll have used all the data without ever observing the same example in more than one stage.

\begin{figure}[t]
    \centering
    \includegraphics[width=1.05\columnwidth]{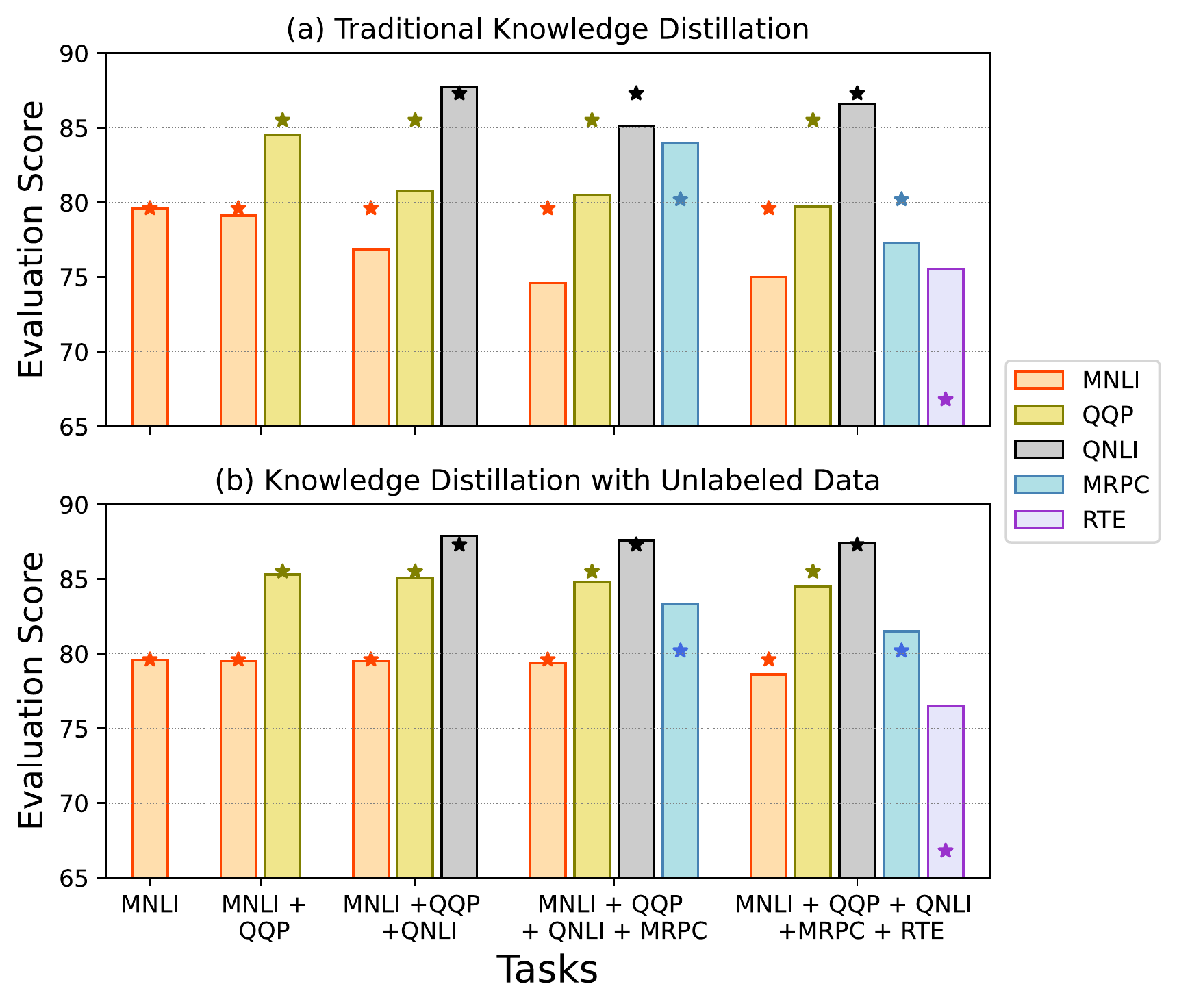}
    \caption{After training an MNLI model, we incrementally add QQP, QNLI, MRPC, and RTE. `\mystar'  markers represent the ST fine-tuning performance for the corresponding task.}
    \label{fig:five_task_chain}
\end{figure}

Figure \ref{fig:five_task_chain} shows that UKD also outperforms TKD in this setting.\footnote{As RTE is smaller than the other datasets, we up-sampled ($4$x) the training data.} Incrementally adding a new task contributes to the forgetting of older tasks for TKD. For example, MNLI performance drops at each step, resulting, at the last stage, in a total drop of about 5\% drop in accuracy (Figure \ref{fig:five_task_chain}a).
However, with UKD we can observe that our approach is able to maintain the performance for old tasks (Figure \ref{fig:five_task_chain}b). For instance, the performance for the first task (MNLI) is almost the same at each step. This experiment demonstrates that the usage of data from the same data distribution of the old tasks is beneficial for avoiding the catastrophic forgetting. Another interesting insight is that when adding new tasks to a model already exposed to several fine-tuning stages, the model is still able to achieve good results on the new task. 
This means that fine-tuning with our strategy does not affect the general knowledge learned in the language model pre-training. Given that the knowledge transfer effect may be limited, as indicated by the TKD results, we argue that the our approach is limiting the loss of the general linguistic knowledge learned in the language model pre-training.
Therefore, our proposed approach is feasible for real-world applications like voice assistants where new tasks with labeled training data could be added to a pool of existing tasks in a model.

In Table \ref{app:tab:results_chain_2} we report the results of an additional task ordering, i.e., QNLI $\rightarrow$ QQP $\rightarrow$ MNLI $\rightarrow$ RTE $\rightarrow$ MRPC. We observe that despite changing the order of the task, the outcome is the same. We observed the similar pattern when we experimented with another task order different than the mentioned ones. Our proposed model is able to limit the catastrophic forgetting happening with other solutions in a continual learning setting.

\begin{table}[!ht]
    \centering
    \begin{tabular}{lcccccc}
         \toprule
            & \textbf{Avg.} & \textbf{QNLI} & \textbf{QQP} & \textbf{MNLI} & \textbf{RTE} & \textbf{MRPC} \\
            & & Acc. & F1 / Acc. & Acc. & Acc. & F1 / Acc.\\
         \midrule
         Single Task & 81.8 & 86.8 & 83.6 / 87.4 & 81.0, 81.6	& 64.3 &	87.9 / 82.1 \\ \midrule
         		
         \multicolumn{7}{l}{Step 2: QQP added to QNLI}\\ \hdashline
         MTL & 85.9 & 87.1 & 83.2 / 87.3 & & & \\
         TKD & 86.1 & 87.2 & 83.5 / 87.7 &  &  & \\
         UKD & 86.4 & 87.7 & 83.6 / 87.9 &  &  &  \\ \midrule
         
         \multicolumn{7}{l}{Step 3: MNLI added to [QNLI, QQP]}\\ \hdashline
         MTL & 83.8 &  87.5 & 83.2 / 87.0 & 80.0, 81.2 & \\
         TKD & 83.4 & 85.9 & 82.6 / 87.0 & 80.3, 81.0 &  &  \\
         UKD & 84.0 & 87.8 & 83.2 / 87.6 & 80.4, 81.1 &  & \\ \midrule
         
         \multicolumn{7}{l}{Step 4: RTE added to [QNLI, QQP, MNLI]}\\ \hdashline

         MTL & 81.9 & 86.8 & 83.4 / 87.2 & 79.6, 80.1 & 74.4 &  \\
         TKD & 80.1 & 84.9 & 81.1 / 85.5 & 75.8, 76.3 & 76.9 &  \\
         UKD & 82.4 & 85.7 & 84.6 / 86.7 & 79.9, 80.0 & 77.2 &   \\

         \multicolumn{7}{l}{Step 5: MRPC added to [QNLI, QQP, MNLI, RTE]}\\ \hdashline
         MTL & 82.0 & 87.4 & 83.1 / 87.0 & 80.3, 81.0 & 74 & 84.8 / 78.7 \\
         TKD & 79.9 & 85.6 & 80.7 / 86.2 & 76.2, 77.3 & 60.3 & 88.7 / 84.1 \\
         UKD & 82.5 & 87.3 & 82.6 / 87.0 & 79.2, 80.2 & 74.7 & 86.9 / 81.9\\
         
         \bottomrule
         
    \end{tabular}
\caption{Results of incrementally learning five tasks. We first fine-tune a pre-trained BERT model on QNLI. Then we incrementally add QQP, MNLI, RTE, and MRPC to that model.}
\label{app:tab:results_chain_2}
\end{table}

\subsection{Continual Learning in Real-World Applications}
\label{sec:realw_exp}

To further validate the assumption that our method can be useful in practical settings, we conduct experiments on a dataset built from traffic directed to a leading digital voice assistant. We tackle the task of Intent Classification with respect to 5 different domains, each containing numerous intents. More specifically, we perform experiments where we add new domains over time. This realistically mimics a practical scenario where voice assistants are dynamically updated to support new functionalities in new domains.

\begin{table}[!ht]
\begin{tabular}{lrrrrr}
\toprule
\textbf{Domain}          & \textbf{Intents} & \multicolumn{1}{l}{\textbf{Train}} & \multicolumn{1}{l}{\textbf{Dev}} & \multicolumn{1}{l}{\textbf{Test}} & \textbf{Skewness}  \\ \midrule
D1  & 12       & {1901}              & {407}             & {408}      & 0.01        \\ 
D2     & 11       & {1311}              & {280}             & {282}      & 2.09        \\ 
D3 & 5        & 995                        & 213                      & 214   & 1.49                    \\ 
D4     & 11       & {611}               & {131}             & {132}         & 2.77     \\ 
D5    & 13       & {396}               & {85}              & {86}          & 2.96     \\ \bottomrule
\end{tabular}%

\caption{Statistics of the voice assistant data. There are multiple intents for each domain and data distribution is generally skewed towards the majority class for most domains.}
\label{tab:intent_data_stat}
\end{table}

\begin{table}
\resizebox{\columnwidth}{!}{%
\begin{tabular}{lrrrrrr}
\toprule
                       & \textbf{D1}     & \textbf{D2}    & \textbf{D3}     & \textbf{D4}    & \textbf{D5}   & \textbf{Avg.} \\ \midrule
Baseline (MTL)    & 0      & 0     & 0      & 0     & 0  & 0    \\
ST & -0.67 &   -1.28 &   -0.79 &   -21.35 &   -30.58 & -10.93\\ \midrule
TKD \textsubscript{[D1,D2]}                & -2.12 & -0.80 &        &    &   &   \textbf{-1.46}    \\
UKD \textsubscript{[D1,D2]}        & -2.31 & -1.93 &        &       &    & -2.12   \\\hdashline
TKD \textsubscript{[D1,D2,D3]}           & -4.89 &   -4.08 &   -3.17 &       & & -4.05       \\
UKD \textsubscript{[D1,D2,D3]}   & -4.05 &   -5.19 &   -1.68 &        &  & \textbf{-3.64}     \\\hdashline
TKD \textsubscript{[D1,D2,D3,D4]}      & -12.9 &   -5.52 &   -5.39 &   -1.55 & & -6.34      \\
UKD \textsubscript{[D1,D2,D3,D4]}       & -3.43 &   -1.59 &   -3.81 &   -5.61 &  & \textbf{-3.61}     \\\hdashline
TKD \textsubscript{[D1,D2,D3,D4,D5]} & -17.74 &   -7.84 &   -7.96 &   -1.34 &   -26.9 & -12.36 \\
UKD \textsubscript{[D1,D2,D3,D4,D5]} & -2.4 &   -0.74 &   -3.95 &   -5.64 &   -3.07 & \textbf{-3.16}\\
\bottomrule
\end{tabular}}%
\caption{Incrementally adding Intent Classification tasks for five different domains. We report the relative F1 score change considering the multi-task setting as the baseline, \ie a negative value indicates a drop from the multi-task baseline. For example, single task fine-tuning on D1 achieves 0.67 less F1 score compared to multi-task training.}
\label{tab:intent_res_l2s}
\end{table}

In Table \ref{tab:intent_data_stat} we report the statistics of the dataset we built with respect to the different domains. The amount of training data in this setting is quite limited, and the number of categories intents for each domain is between 5 and 13.

In Table \ref{tab:intent_res_l2s} we show the results of the Intent Classification on the five different domains by incrementally adding one domain to a pre-trained large language model. Again, we adopted the BERT-base-uncased model, fine-tuned for 20 epochs with early stopping. We adopted the same hyper-parameters as in the previous experiments, except for the batch size. Given the limited data, we adopted a batch size of 16. For reasons of confidentiality, we report the relative performance compared to the multi-task setting. As in the previous experiments, the MTL performance can be considered as an upper bound. The results show that every task gets little to large performance gain by the traditional multi-task training with respect to a ST fine-tuning (first two rows of Table \ref{tab:intent_res_l2s}). This is intuitive as in the MTL training the model can exploit information shared across tasks. The gain is high for the domains with smaller datasets. For example, D5, which is the least represented domain, gets a boost of 30 points with respect to ST.

In consecutive rows, we observe that, when we add D2 to a model trained only on D1, TKD performs slightly better than UKD. However, when we add more tasks, UKD starts outperforming TKD. For example, adding D3 and D4 causes TKD to  drop of about 12.9 and 5.5 points with respect to the first task and second domains. Instead, adding D3 with UKD results in a drop of only 3.43 and 1.59 for the two domains, respectively. Sometimes the performance of the newest task is better of TKD than UKD (e.g., when adding D4, TKD drop with respect to MTL is 1.55 while UKD drop is 5.61). As we have the control over the data of newly added tasks, we can apply techniques like oversampling or hyper-parameter tuning to improve the new task's performance. In general, the results obtained with UKD seems more promising, as the average drop over all the steps for all the domains with respect to the MTL baseline is -3.42 for UKD while it is -7.3 for TKD.

\section{Conclusion}\label{sec:conclusions}
In this paper, we proposed a semi-supervised approach based on the Knowledge Distillation framework to incrementally learn new natural language tasks in already trained models. We concentrated our analysis on maintaining a consistent performance on the tasks already learned, \ie to prevent the catastrophic forgetting of the model. We suggest the usage of an unlabeled set of data coming from the same data distribution of the previously acquired tasks to stabilize the training in the knowledge distillation framework. Experimental results on both publicly available benchmarks and a dataset we built from real voice-assistant data demonstrate that the usage of unlabeled data from previous task distribution is crucial to prevent the catastrophic forgetting phenomenon when dealing with different data distributions.

We expect this technique to be useful for many practical problems when a model must be updated over time to support new tasks. This is a common scenario in industry settings, where systems are updated over time to support new functionalities. In the future, it may be interesting to study the problem of automatically generating data representing the old task data distributions starting from the already trained models.

\bibliography{biblio,anthology}
\bibliographystyle{ACM-Reference-Format}

\end{document}